\pdfoutput=1

\documentclass[11pt]{article}

\usepackage{acl}

\usepackage{times}
\usepackage{latexsym}
\usepackage[T1]{fontenc}

\usepackage[utf8]{inputenc}

\usepackage{microtype}

\usepackage{amsmath}
\usepackage{booktabs}
\usepackage{multirow}
\usepackage{graphicx}

\usepackage{CJKutf8}
\usepackage{color}
\usepackage{arydshln}

\title{
Effective Token Graph Modeling using a Novel Labeling Strategy for Structured Sentiment Analysis
}

\author{Wenxuan Shi, Fei Li, Jingye Li, Hao Fei, Donghong Ji\textsuperscript{$\dagger$} \\
         Key Laboratory of Aerospace Information Security and Trusted Computing, Ministry of \\
Education, School of Cyber Science and Engineering, Wuhan University, Wuhan, China \\  \texttt{\{shiwenxuan,lifei\_csnlp,theodorelee,hao.fei,dhji\}@whu.edu.cn}}

\begin{document}
\begin{CJK}{UTF8}{gbsn}
\maketitle
\renewcommand{\thefootnote}{\fnsymbol{footnote}}
\footnotetext[2]{Corresponding author}
\renewcommand{\thefootnote}{\arabic{footnote}}
\begin{abstract}


The state-of-the-art model for structured sentiment analysis casts the task as a dependency parsing problem, which has some limitations: (1) The label proportions for span prediction and span relation prediction are imbalanced. (2) The span lengths of sentiment tuple components may be very large in this task, which will further exacerbates the imbalance problem. (3) Two nodes in a dependency graph cannot have multiple arcs, therefore some overlapped sentiment tuples cannot be recognized.
In this work, we propose nichetargeting solutions for these issues. First, we introduce a novel labeling strategy, which contains two sets of token pair labels, namely essential label set and whole label set.
The essential label set consists of the basic labels for this task, which are relatively balanced and applied in the prediction layer. 
The whole label set includes rich labels to help our model capture various token relations, which are applied in the hidden layer to softly influence our model.
Moreover, we also propose an effective model to well collaborate with our labeling strategy, which is equipped with the graph attention networks to iteratively refine token representations, and the adaptive multi-label classifier to dynamically predict multiple relations between token pairs.
We perform extensive experiments on 5 benchmark datasets in four languages.
Experimental results show that our model outperforms previous SOTA models by a large margin.\footnote{Our code is available at \href{https://github.com/Xgswlg/TGLS}{\texttt{https://github.com/\newline Xgswlg/TGLS}}}


\end{abstract}

\section{Introduction}
\begin{figure}[t]
\centering
\includegraphics[width=1\columnwidth]{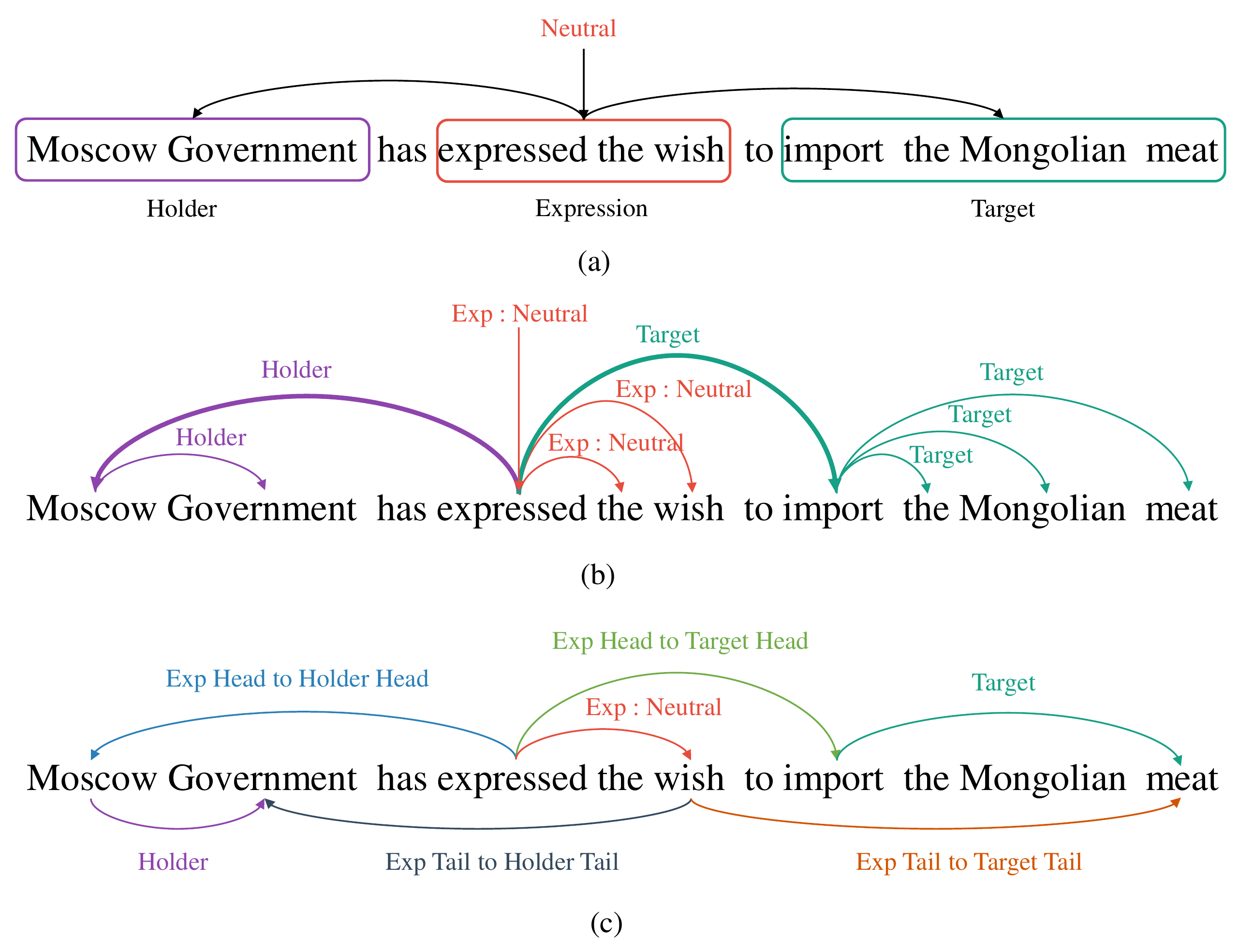} 
\caption{
\textbf{(a)} An example of structured sentiment analysis. 
\textbf{(b)} The head-first parsing graph proposed by \citet{structured}, where the arcs related to holder(target)-expression linking relations are bold.
\textbf{(c)} Our proposed \textit{essential label set}, which has more balanced label distribution for holder, target or expression span prediction and their linking relation prediction.
}
\label{fig1}
\end{figure}

Structured Sentiment Analysis (SSA), which aims to predict a structured sentiment graph as shown in Figure \ref{fig1}(a), can be formulated into the problem of tuple extraction, where a tuple $(h,e,t,p)$ denotes a holder $h$ who expressed an expression $e$ towards a target $t$ with a polarity $p$.
SSA is a more challenging task, because other related tasks only focus on extracting part of tuple components or the text spans of the components are short.
For example, \textit{Opinion Role Labeling} \citep{katiyar2016investigating,xia2021unified} does not include the extraction of sentiment polarities,
and \textit{Aspect-Based Sentiment Analysis} (ABSA) \citep{absa,wang-absa} extracts the aspect and opinion terms typically consisting of one or two words.
The state-of-the-art SSA model is proposed by \newcite{structured}, which casts the SSA task as the dependency parsing problem and predicts all tuple components as a dependency graph (Figure \ref{fig1}(b)).

\begin{figure*}[t]
\centering
\includegraphics[width=2.1\columnwidth]{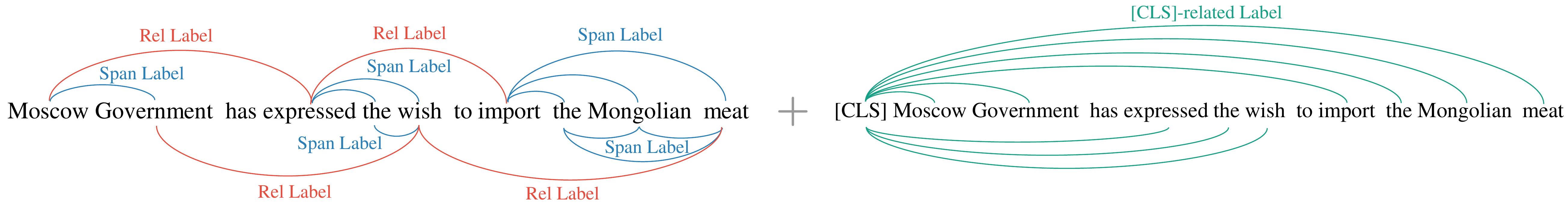} 
\caption{The whole label set contains the labels for span prediction and span relation prediction, as well as the \texttt{[CLS]}-related labels that connect a sentinel \texttt{[CLS]} token with the holder, target and expression tokens.}
\label{fig2}
\end{figure*}

However, their method exists some shortages. 
Taking Figure \ref{fig1}(b) as example, only 2 arcs (e.g., expressed$\rightarrow$import and expressed$\rightarrow$Moscow) are related to span linking relation prediction (i.e., the relations between expressions and holders or targets), 
while much more other arcs are related to 
span prediction (e.g., import$\rightarrow$the and import$\rightarrow$meat).
Such imbalanced labeling strategy will make the model pay more attention on span prediction but less on span relation  prediction.
Furthermore, since the span lengths of sentiment tuple components may be very large in the SSA task,
the label imbalanced problem will become more severe.
Besides, the dependency parsing graph is not able to deal with multi-label classification, since it does not allow multiple arcs to share the same head and dependent tokens. Therefore, some overlapped sentiment tuples cannot be recognized. The statistics of span length and multi-label problems are listed in Table \ref{multi-label}.

\begin{table}[t]
\footnotesize
\renewcommand{\arraystretch}{1}
\centering
\begin{tabular}{@{}lrrrr@{}}
\toprule
\multicolumn{1}{c}{\multirow{2}{*}{Dataset}} & \multicolumn{3}{c}{Span Length $\geq$ 4}   &  \multirow{2}{*}{\begin{tabular}[c]{@{}c@{}}Multi\\ Label\end{tabular}}  \\ \cmidrule(l){2-4} 
\multicolumn{1}{c}{}   & \multicolumn{1}{c}{Hoder} & \multicolumn{1}{c}{Target} & \multicolumn{1}{c}{Exp.} & \multicolumn{1}{c}{}  \\ \midrule
\textbf{NoReC}$_\textsuperscript{Fine}$      & 1.1\%                     & 19.2\%                     & 56.8\%            & 14.0\%              \\
\textbf{MultiB}$_\textsuperscript{CA}$                                                                                     & 2.6\%                     & 18.4\%                     & 21.4\%           & 8.7\%               \\
\textbf{MultiB}$_\textsuperscript{EU}$                                                                                    & 1.1\%                     & 2.7\%                      & 15.3\%                & 3.6\%           \\
\textbf{MPQA}                                                                                   & 19.9\%                    & 51.1\%                     & 14.5\%         & 1.0\%                 \\
\textbf{DS}$_\textsuperscript{Unis}$                                                                                   & 1.3\%                     & 0.8\%                      & 13.7\%         & 1.9\%                   \\ \bottomrule
\end{tabular}
\caption{Statistics of the proportion of each sentiment component whose span length (in tokens) is greater than or equal to 4, and the proportion of sentences requiring multi-label classification for SSA. 
}
\label{multi-label}
\end{table}

To alleviate the label imbalance problem in \citet{structured},
we propose a novel labeling strategy that consists of two parts: First, we design a set of labels called \textbf{essential label set} (Figure \ref{fig1}(c)), which can be considered as the basic label set for decoding SSA tuples, since it only includes the labels to tag the boundary tokens of spans. As seen, the proportion of span prediction labels and span relation prediction labels are relatively balanced, so that we can mitigate the label imbalance problem and meanwhile keep the basic ability of extracting sentiment tuples if the essential label set is learnt in the final prediction layer of our model.


However, the labels related to recognize non-boundary tokens of SSA components are also important. For instance, they can encode the relations between the tokens inside the spans, which may benefit the extraction of the holders, expressions or targets with long text spans.
To this end, we design another label set called \textbf{whole label set} (Figure \ref{fig2}), which includes richer labels to fully utilize various information such as the relations among boundary tokens, non-boundary tokens, the tokens within a span, the tokens across different spans.
Moreover, since the dependency-based method \cite{structured} only considers the local relation between each pair of tokens, we add the labels between \texttt{[CLS]} and other tokens related to sentiment tuples into our whole label set, in order to utilize sentence-level global information.
Considering that if the whole label set is directly applied on the output label for training, the label imbalance problem may occur again. We instead employ the whole label set in a soft and implicit fashion by applying it on the hidden layer of our model. 

To well collaborate with our labeling strategy, we also propose an effective token graph model, namely \textbf{TGLS} (\textbf{T}oken \textbf{G}raph with a novel \textbf{L}abeling \textbf{S}trategy), which uses rich features such as word, part-of-speech tags and characters as inputs and yields contextualized word representations by BiLSTM and multilingual BERT\cite{bert}.
In the hidden layer, we build a multi-view token graph, which has four views corresponding to different relations in the whole label set and each view is a graph attention network \cite{gat} with token representations as the nodes. 
In the prediction layer, we introduce a novel adaptive multi-label classifier to extract all the sentiment tuples no matter that they are overlapped or not.

We conduct extensive experiments on five benchmarks, including \text{NoReC}$_\textsuperscript{Fine}$ \citep{norec}, \text{MultiB}$_\textsuperscript{EU}$,  \text{MultiB}$_\textsuperscript{CA}$ \citep{euca}, \text{MPQA} \citep{Wiebe2005Annotating} and \text{DS}$_\textsuperscript{Unis}$ \citep{ds}.
The resluts show that our TGLS model outperforms the SOTA model by a large margin.
In summary, our main contributions include:
\begin{itemize}
\setlength{\itemsep}{0pt}
\setlength{\parsep}{0pt}
\setlength{\parskip}{0pt}
\item We design a novel labeling strategy to address the label imbalance issue in prior work. Concretely, we employ the whole label set and essential label set in the hidden and prediction layer respectively, achieving a balance between the label variety and label imbalance.


\item We propose an effective token graph model to well collaborate with our labeling strategy, which learns the token-token relations via multi-view token graph networks and reasons the labels between each pair of words using the adaptive multi-label classifier for both overlapped and non-overlapped tuple extraction.


\item The experimental results show that our model has achieved the SOTA performance in 5 datasets for structured sentiment analysis, especially in terms of the end-to-end sentiment tuple extraction.

\end{itemize}

\begin{figure*}[t]
\centering
\includegraphics[width=1\textwidth]{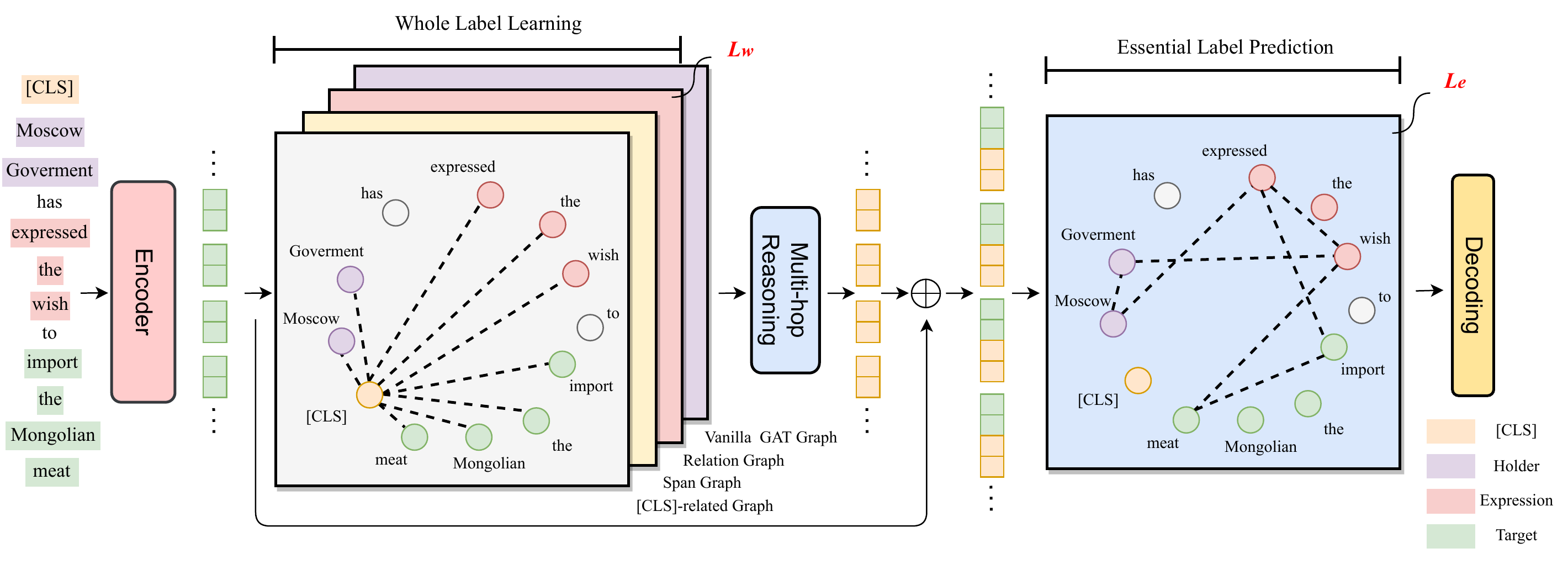} 
\vspace{-0.6cm}
\caption{Overall architecture of the our framework. From left to right, the first is an encoder to yield contextualized word representations from input sentences, and the next is a graph layer where we produce attention scoring matrices by whole label prediction. Then we build a multi-hop reasoning layer and refine token representations. 
Finally, a prediction layer is leveraged for reasoning the relations in essential labels and based on which we decode all components of an opinion tuple.}
\label{fig3}
\end{figure*}

\section{Related Work}
The task of the Structured Sentiment Analysis (SSA) can be divided into sub-tasks such as span extraction of the holder, target and expression, relation prediction between these elements and assigning polarity. Some existing works in \textit{Opinion Mining} used pipeline methods to first extract spans and then the relations mostly on the MPQA dataset \citep{Wiebe2005Annotating}. 
For example, \citet{katiyar2016investigating} propose a BiLSTM-CRF model which is the first such attempt using a deep learning approach, \citet{zhang2019end} propose a transition-based model which identifies opinion elements by the human-designed transition actions, and \citet{xia2021unified} propose a unified span-based model to jointly extract the span and relations. However, all of these works ignore the polarity classification sub-task.

In \textit{End2End Aspect-Based Sentiment Analysis} (ABSA), there are also some attempts to unify several sub-tasks. For instance, \citet{wang-absa} augment the ABSA datasets with sentiment expressions, \citet{IMN} make use of this data and models the joint relations between several sub-tasks to learn common features, and \cite{racl} also exploit interactive information from each pair of sub-tasks (target extraction, expression extraction, sentiment classification).
However, \citet{wang-absa} only annotate sentiment-bearing words not phrases and do not specify the relationship between target and expression, it therefore may not be adequate for full structured sentiment analysis.

Thus, \citet{structured} propose a unified approach in which they formulate the structured sentiment analysis task into a dependency graph parsing task and jointly predicts all components of a sentiment graph. However, as aforementioned, this direct transformation may be problematic as it may introduce label imbalance in span and relation prediction. Thus, we propose an effective graph model with a novel labeling strategy in which we employ a whole label set in the hidden layer to softly affect our model, and an essential label set in the prediction layer to address the imbalance issue.

The design of our essential label set is inspired by the Handshaking Tagging Scheme \citep{tplinker}, which is a token pair tagging scheme for entity and relation extraction. The handshaking tagging scheme involves only the labels related to the boundary tokens and enables a one-stage joint extraction of spans and relations. In our work, we modify the handshaking tagging scheme to use it for SSA. Furthermore, since the component span of this task is relatively long, only utilizing the boundary tokens cannot make full use of the annotation information, so we propose a new label set called whole label set, which together with essential label set constitutes our labeling strategy.

\section{Token-Pair Labeling Strategy}


\subsection{Essential Label Set}
Our essential label set only involves the labels related to the boundary tokens, therefore the label proportions for span prediction and span relation prediction are relatively balanced.
Given a sentence "\textit{Moscow government has expressed the wish to import the Mongolian meat.}", the essential label set consists of the following labels:
\begin{itemize}
\setlength{\itemsep}{0pt}
\setlength{\parsep}{0pt}
\setlength{\parskip}{0pt}
\item \textit{Holder}: \textit{Moscow} $\rightarrow$ \textit{government}
\item \textit{Exp:Neutral}: \textit{expressed} $\rightarrow$ \textit{Moscow}
\item \textit{Target}: \textit{import} $\rightarrow$ \textit{meat}
\item \textit{Exp Head to Holder Head}: \textit{expressed} $\rightarrow$ \textit{Moscow}
\item \textit{Exp Tail to Holder Tail}: \textit{wish} $\rightarrow$ \textit{government}
\item \textit{Exp Head to Target Head}: \textit{expressed} $\rightarrow$ \textit{import}
\item \textit{Exp Tail to Target Tail}: \textit{wish} $\rightarrow$ \textit{meat}
\end{itemize}
where the \textit{Holder}, \textit{Exp.} and \textit{Target} represent the three components of a sentiment tuple, the \textit{Head} or \textit{Tail} means the start or end token of a component, and the \textit{Neutral} denotes the polarity.


\subsection{Whole Label Set}

Our whole label set involves both the labels related to boundary and non-boundary tokens, as well as the labels related to \texttt{[CLS]} and all tokens in the sentiment tuples. Thus, our whole label set can be divided into three groups, span labels, relation labels and \texttt{[CLS]}-related labels.
Given the sentence in Figure \ref{fig2}, the whole label set include the following labels:
\begin{itemize}
\setlength{\itemsep}{0pt}
\setlength{\parsep}{0pt}
\setlength{\parskip}{0pt}
\item \textit{Span Label}: e.g. \textit{import} $\rightarrow$ \textit{Mongolian}
\item \textit{Rel Label}: e.g. \textit{Moscow} $\rightarrow$ \textit{expressed}
\item \textit{\texttt{[CLS]}-related Label}: e.g. \textit{\texttt{[CLS]}} $\rightarrow$ \textit{expressed}
\end{itemize}
where the span and relation labels make our model be aware of the token relations inside and across the spans of sentiment components, and \texttt{[CLS]}-related labels can help our model to capture the sentence-level global information. 
We apply whole labels in the hidden layer to softly embed the above information into our model, in order to avoid the potential label imbalance issue.


\subsection{Decoding}
We first decode all the expression-holder and expression-target pairs that meet the constraints of essential label set. In detail, we can get all component spans based on span prediction labels (e.g. \textit{Holder}, \textit{Exp:Neutral} and \textit{Target} labels), then we decode all expression to holder or target pairs as long as it meets one of the corresponding relation prediction labels (e.g. for expression to holder pairs, the labels are \textit{Exp Head to Holder Head} and \textit{Exp Tail to Hoder Tail}). After decoding all the component pairs, we enumerate all possible triples from pairs with the same expression, thus finally decode all the sentiment tuples.

\section{Methodology}
In this section, We formally present our proposed TGLS model in detail (Figure \ref{fig3}), which mainly consists of four parts, the encoder layer, the multi-view token graph as the hidden layer, the adaptive multi-label classifier as the prediction layer and the hierarchical learning strategy to train our model.


\subsection{Encoder Layer}
Consider the $i^{th}$ token in a sentence with $n$ tokens, we represent it by concatenating its
token embedding $\mathbf{e}_{i}^{word}$, part-of-speech (POS) embedding $\mathbf{e}_{i}^{pos}$, lemma embedding $\mathbf{e}_{i}^{lemma}$, and character-level embedding $\mathbf{e}_{i}^{char}$ together:
\begin{align}
    \boldsymbol{w}_{i}=\mathbf{e}_{i}^{word}\oplus\mathbf{e}_{i}^{pos}\oplus\mathbf{e}_{i}^{lemma}\oplus\mathbf{e}_{i}^{char}
\end{align}
where $\oplus$ denotes the concatenation operation. 
The character-level embedding is generated by the convolution neural networks (CNN) \citep{charcnn}.
Then, we employ bi-directional LSTM (BiLSTM) to encode the vectorial token representations into contextualized word representations:
\begin{align}
\boldsymbol{h}_{i} =\operatorname{BiLSTM}\left(\boldsymbol{w}_{i}\right)
\end{align}
where $\boldsymbol{h}_{i}$ is the token hidden representation.

Moreover, in the same way as previous work \citep{structured}, we also enhance token representations with pretrained contextualized embeddings using multilingual BERT \citep{bert}.

\subsection{Multi-view Token Graph}
\label{one-to-one}
In this section, we propose a novel multi-view token graph as our hidden layer, which includes four views, span graph, relation graph, \texttt{[CLS]}-related graph and vanilla GAT graph, and each view is full connected with the attention scoring weights as graph edges and the token representations as graph nodes. Recall that the whole label set is applied in this layer, which includes three groups of labels (span, relation and \texttt{[CLS]}-related labels). Thus, three views of graphs (span, relation and \texttt{[CLS]}-related graph) are used to digest information from three groups of labels respectively, while one view (vanilla GAT graph) is not assigned for any specific task, as the method in vanilla graph attention network (GAT) \cite{gat}. 
Formally, we represent the latent token graph $\boldsymbol{G}$ as follows:
\begin{align}
\begin{split}
&\boldsymbol{G}=\left({\textbf{V}},{S^{\mathcal{G}}_{o},S^{\mathcal{G}}_{s},S^{\mathcal{G}}_{r},S^{\mathcal{G}}_{c}}\right) \\
\end{split}
\end{align}
where superscript $\mathcal{G}$ denotes the graph layer, ${\textbf{V}}$ is the set of tokens, $S^{\mathcal{G}}_{o}$ is the attention scoring matrix in vanilla GAT, 
$S^{\mathcal{G}}_{s}$, $S^{\mathcal{G}}_{r}$ and $S^{\mathcal{G}}_{c}$
are the attention scoring matrices used to capture information from span, relation and \texttt{[CLS]}-related labels respectively. Without loss of generality, we employ $\mathcal{S}^{\mathcal{G}}=\{S^{\mathcal{G}}_{o}, S^{\mathcal{G}}_{s}, S^{\mathcal{G}}_{r}, S^{\mathcal{G}}_{c}\}$ unifiedly to represent the four matrices.

\subsubsection{Graph Induction}
In this section, we introduce the process that we induce the edges of our multi-view token graphs (i.e. four attention scoring matrices $\mathcal{S}^{\mathcal{G}}$) using a mechanism of attention scoring.

\paragraph{Attention Scoring}  Our attention matrices are produced by a mechanism of attention scoring which takes two token representations $\boldsymbol{h}_{i},\boldsymbol{h}_{j}$ as the input, and for the attention matrix corresponding to a certain view $v\in\{o,s,r,c\}$,
we first map the tokens to ${q}_{v,i}$ and ${k}_{v,j}$ with two multi-layer perceptions (MLP):
\begin{align}
\begin{split}
{q}_{v,i},{k}_{v,j}={MLP}^{q}_{v}\left(\boldsymbol{h}_{i}\right),{MLP}^{k}_{v}\left(\boldsymbol{h}_{j}\right)
\end{split}
\label{eq4}
\end{align}
Then we apply the technique of Rotary Position Embedding (RoPE) \cite{RoFormer} to encode the relative position information. Thus, for the graph of view $v$, the attention score $S^{\mathcal{G}}_{v,ij}$ between token $i$ and $j$ can be calculated as follows:
\begin{align}
\begin{split}
    S^{\mathcal{G}}_{v,ij} =\left({q}_{v,i}\right)^\top \boldsymbol{R}_{j-i} {k}_{v,j}
\end{split}
\label{eq5}
\end{align}
where $\boldsymbol{R}_{j-i}$ can incorporate explicit relative positional information into the attention score $S^{\mathcal{G}}_{v,ij}$. And in the same way as calculating $S_{v,ij}^\mathcal{G}$, we can produce the scores of all views and all token pairs, thus inducing the whole graph edges $\mathcal{S}^{\mathcal{G}}$:
\begin{align}
    \mathcal{S}^{\mathcal{G}} = \left\{S^{\mathcal{G}}_{v,ij}|v\in\{o,s,r,c\},1\leq i,j\leq n \right\}
\label{eq6}
\end{align}
where $n$ is the length of the sentence.
The process that the whole label set learnt by attention scoring matrices $S^{\mathcal{G}}_{s}$, $S^{\mathcal{G}}_{r}$ and $S^{\mathcal{G}}_{c}$ through a multi-label adaptive-threshold loss will be introduced in Section \ref{sec:hierarchical learning}.

\subsubsection{Multi-hop Reasoning}
Considering that the attention scoring matrix $\mathcal{S}^\mathcal{G}$ now fuses rich information, we naturally think of applying a multi-hop reasoning to obtain more informative token representations. Concretely, we first apply a softmax on our adjacency attention matrix $\mathcal{S}^{\mathcal{G}}$, then the computation for the representation $\boldsymbol{u}_{i}^{l+1}$ of the token $i$ at the $(l+1)^{th}$layer, which takes the representations from previous layer as input and outputs the updated representations, can be defined as:
\begin{align}
A_{v} &= Softmax\left(S^\mathcal{G}_{v}\right), v\in\{o,s,r,c\} \\
\boldsymbol{u}_{i}^{l+1}&=\sigma\left(\frac{1}{N} \sum_{v} \sum_{j \in \mathcal{N}_{i}^{v}} A_{v,ij} \boldsymbol{W}^{v}_{l} \boldsymbol{u}_{j}^{l}\right)
\end{align}
where $\boldsymbol{W}^{v}_{l}$ is the trainable weight, $\mathcal{N}_{i}^{v}$ is the neighbor of token $i$ in graph of view $v$, $\sigma$ is the ReLU activation function.


\subsection{Adaptive Multi-label Classifier}
Considering that the previous sota model \cite{structured} is not able to deal with multi-label classification as aforementioned, we propose a novel adaptive multi-label classifier as our prediction layer to identify possible essential labels for each token pair. 

Firstly, we take a shortcut connection between the outputs of the encoder layer and graph layer to get the final representation $\boldsymbol{c}_{i} = \boldsymbol{h}_{i}\oplus\boldsymbol{u}_{i}$ for each token.
And by taking $\boldsymbol{c}_{i}$ as the input, we calculate the attention scoring matrices $\mathcal{S}^\mathcal{P}$ based on the mechanism of attention scoring (cf. Eq.(\ref{eq4}), Eq.(\ref{eq5}) and Eq.(\ref{eq6})):
\begin{align}
    \mathcal{S}^\mathcal{P}=\{S_{r}^\mathcal{P}| r\in{\mathcal{R}_{e}}\}
\end{align}
where superscript $\mathcal{P}$ denotes the prediciton layer, $\mathcal{R}_{e}$ denotes the essential label set.
Then, we introduce a technique of adaptive thresholding, which produces a token pair dependent threshold to enable the prediction of the labels for each token pair.

\paragraph{Adaptive Thresholding}
For a certain token pair with representations of $\boldsymbol{c}_{i},\boldsymbol{c}_{j}$, the token pair dependent threshold $TH^{\mathcal{P}}_{ij}$ and the whole $TH^{\mathcal{P}}$ are calculated as follows:
\begin{align}
\begin{split}
TH_{ij}^\mathcal{P}& = \left({q}^{TH}_{i}\right)^\top \boldsymbol{R}_{j-i} \boldsymbol{k}^{TH}_{j} \\
TH^{\mathcal{P}} &= \left\{TH_{ij}^\mathcal{P}|1\leq i,j\leq n \right\}
\end{split}
\label{eq9}
\end{align}
where ${q}^{TH}_{i}=\boldsymbol{W}_{q}\boldsymbol{h}_{i}+\boldsymbol{b}_{q}, {k}^{TH}_{j}=\boldsymbol{W}_{k}\boldsymbol{h}_{j}+\boldsymbol{b}_{k}$, the $\boldsymbol{W}_{q}$, $\boldsymbol{W}_{k}$, $\boldsymbol{b}_{q}$ and $\boldsymbol{b}_{k}$ are the trainable weight and bias matrix, $\boldsymbol{R}_{j-i}$ are calculated in the same way as Eq.(\ref{eq5}), which is used to incorporate explicit relative positional information. 

Formally, for a certain token pair $c_i,c_j$, the essential label set is predicted by the following equation:
\begin{align}
\Omega_{ij}=\left\{ r|S_{r,ij}^\mathcal{P}>TH_{ij}^{\mathcal{P}}, r\in{\mathcal{R}_{e}}\right\}
\end{align}
where $\mathcal{R}_{e}$ denotes the essential label set, $\Omega_{ij}$ is the set of predicted labels of token pair $c_i,c_j$.


\subsection{Training}
\label{sec:hierarchical learning}
In this section, we will propose a novel loss function, namely multi-label adaptive-threshold loss, to enable a hierarchical training process for our model and our labeling strategy (i.e. whole label set learnt by $S^{\mathcal{G}}_{s}$, $S^{\mathcal{G}}_{r}$ and $S^{\mathcal{G}}_{c}$ in the hidden layer, essential label set learnt by $\mathcal{S}^{\mathcal{P}}$ in the prediction layer), which is based on a variant\footnote{The variant of Circle loss was proposed by Su on the website \href{https://kexue.fm/archives/7359}{https://kexue.fm/archives/7359}.} of Circle loss \cite{circleloss}, the difference is that we replace the fixed global threshold with the adaptive token pair dependent threshold to enable a flexible and selective learning of more useful information from whole label set.

Take the hidden layer as an example. Actually, we also implement the adaptive thresholding (cf. Eq.(\ref{eq9})) in the hidden layer, where we compute all the token pair dependent threshold $TH^{\mathcal{G}} = \left\{TH_{ij}^\mathcal{G}|1\leq i,j\leq n \right\}$ by taking the token representation $\boldsymbol{h}_{i}$ and $\boldsymbol{h}_{j}$ as the input.
Then, the multi-label adaptive-threshold loss in hidden layer can be calculated as follows:
\begin{align}
\begin{split}
    \mathcal{L}_{w}&=\sum_{i}\sum_{j>i}\log\left(e^{TH^{\mathcal{G}}_{ij}}+\sum_{r \in \Omega^{neg}_{ij}} e^{S^{\mathcal{G}}_{r,ij}}\right)\\
&+\sum_{i}\sum_{j>i}\log \left(e^{-TH^{\mathcal{G}}_{ij}}+\sum_{r \in \Omega^{pos}_{ij}} e^{-S^{\mathcal{G}}_{r,ij}}\right)
\end{split}
\end{align}
where $\Omega^{pos}_{ij}\subseteq{\mathcal{R}_{w}}$ and $\Omega^{neg}_{ij}\subseteq{\mathcal{R}_{w}}$ are positive and negative classes involving whole labels that exist or not exist between token $i$ and $j$.
When minimizing $\mathcal{L}_{w}$, the loss pushes the attention score $S_{r,ij}^\mathcal{G}$ above the threshold $TH_{ij}^\mathcal{G}$ if the token pair possesses the label, while pulls below when it does not.\footnote{As aforementioned in Section \ref{one-to-one}, three of the attention scoring matrices and three groups of the whole labels have a one-to-one relationship, so here we can index the three matrices with the whole labels.}

In a similar way we can calculate the loss $\mathcal{L}_{e}$ in the prediction layer by taking $TH^{\mathcal{P}}$,  $\mathcal{S}^{\mathcal{P}}$ as the inputs of the loss function.
Thus the whole loss of our model can be calculated as follows:
\begin{align}
    \mathcal{L}_{all} =& \mathcal{L}_{e} + \alpha\mathcal{L}_{w}
\label{alpha}
\end{align}
where the $\alpha$ is a hyperparameter to adjust the ratio of the two losses.

\section{Experiments}

\begin{table*}[t]
\centering
\renewcommand{\arraystretch}{0.9}
\resizebox{1.8\columnwidth}{!}{
\begin{tabular}{@{}llccccccc@{}}
\toprule
\multirow{2}{*}{Dataset}       & \multirow{2}{*}{Model} & \multicolumn{4}{c}{Span}                      & Targeted      & \multicolumn{2}{c}{Sent. Graph} \\ \cmidrule(l){3-6}\cmidrule(l){7-7}\cmidrule(l){8-9} 
                               &                        & Holder F1     & Target F1     & Exp. F1 & Overall F1      & F1            & NSF1           & SF1   \\ \midrule
\multirow{4}{*}{\textbf{NoReC}$_\textsuperscript{Fine}$}         & RACL-BERT              & -             & 47.2          & 56.3 &   -        & 30.3          & -              & -              \\  
                               & Head-first             & 51.1          & 50.1          & 54.4  &  53.1$^*$      & 30.5          & 37.0             & 29.5           \\ 
                               & Head-final             & 60.4          & \textbf{54.8} & 55.5  & 55.7$^*$       & 31.9          & 39.2           & 31.2           \\\cmidrule(l){3-6}\cmidrule(l){7-7}\cmidrule(l){8-9}
                               & TGLS            & \textbf{60.9} & 53.2          & \textbf{61.0} & \textbf{58.1}\textcolor{white}{$^*$} & \textbf{38.1} & \textbf{46.4}  & \textbf{37.6}  \\ \midrule
\multirow{4}{*}{\textbf{MultiB}$_\textsuperscript{EU}$}            & RACL-BERT              & -             & 59.9          & 72.6  & -       & 56.8          & -              & -              \\ 
                               & Head-first             & 60.4          & 64.0            & 73.9  & 69.6$^*$ & 57.8          & 58.0             & 54.7           \\ 
                               & Head-final             & 60.5          & 64.0            & 72.1 &  68.2$^*$       & 56.9          & 58.0             & 54.7           \\\cmidrule(l){3-6}\cmidrule(l){7-7}\cmidrule(l){8-9}
                               & TGLS            & \textbf{62.8} & \textbf{65.6} & \textbf{75.2} & \textbf{71.0}\textcolor{white}{$^*$} & \textbf{60.9} & \textbf{61.1}  & \textbf{58.9}  \\ \midrule
\multirow{4}{*}{{\textbf{MultiB}}$_\textsuperscript{CA}$}            & RACL-BERT              & -             & 67.5          & 70.3  &  -      & 52.4          & -              & -              \\ 
                               & Head-first             & 43.0            & 72.5          & 71.1 & 70.5$^*$ & 55.0            & 62.0             & 56.8           \\ 
                               & Head-final             & 37.1          & 71.2          & 67.1  &   70.2$^*$     & 53.9          & 59.7           & 53.7           \\\cmidrule(l){3-6}\cmidrule(l){7-7}\cmidrule(l){8-9}
                               & TGLS            & \textbf{47.4} & \textbf{73.8} & \textbf{71.8} & \textbf{71.6}\textcolor{white}{$^*$}& \textbf{60.6} & \textbf{64.2}  & \textbf{59.8}  \\ \midrule
\multirow{4}{*}{\textbf{MPQA}} & RACL-BERT              & -             & 20.0            & 31.2  & -        & 17.8          & -              & -              \\ 
                               & Head-first             & 43.8          & 51.0            & \textbf{48.1} & \textbf{47.7}$^*$ & \textbf{33.5} & 24.5           & 17.4           \\  
                               & Head-final             & \textbf{46.3} & 49.5          & 46.0  &  47.2$^*$        & 18.6          & 26.1           & 18.8           \\\cmidrule(l){3-6}\cmidrule(l){7-7}\cmidrule(l){8-9}
                               & TGLS            & 44.1          & \textbf{51.7} & 47.8 & 47.0\textcolor{white}{$^*$}& 23.3          & \textbf{28.2}  & \textbf{21.6}  \\ \midrule
\multirow{4}{*}{\textbf{DS}$_\textsuperscript{Unis}$}            & RACL-BERT              & -             & 44.6          & 38.2  &   -     & 27.3          & -              & -              \\ 
                               & Head-first             & 28.0            & 39.9          & 40.3   &    40.1$^*$   & 26.7          & 31.0             & 25.0             \\  
                               & Head-final             & 37.4          & 42.1          & \textbf{45.5} & 43.0$^*$ & 29.6          & 34.3           & 26.5           \\ \cmidrule(l){3-6}\cmidrule(l){7-7}\cmidrule(l){8-9}
                               & TGLS          & \textbf{43.7} & \textbf{49.0} & 42.6  &    \textbf{45.7}\textcolor{white}{$^*$}    & \textbf{31.6} & \textbf{36.1}  & \textbf{31.1}  \\ \bottomrule
\end{tabular}}
\caption{Main experimental results of our TGLS model and comparison with previous works. The score marked as bold means the best performance among all the methods.
The baseline results with "$^*$" are from our reimplementation, the others are from \citep{structured}.
}
\label{table1}
\end{table*}

\subsection{Datasets and Configuration}
For comparison with previous sota work \citep{structured}, we perform experiments on five structured sentiment datasets in four languages, including multi-domain professional reviews \textbf{NoReC}$_\textsuperscript{Fine}$ \citep{norec} in Norwegian, hotel reviews \textbf{MultiB}$_\textsuperscript{EU}$ and \textbf{MultiB}$_\textsuperscript{CA}$ \citep{euca} in Basque and Catalan respectively, news \textbf{MPQA} \citep{Wiebe2005Annotating} in English and reviews of online universities and e-commerce \textbf{DS}$_\textsuperscript{Unis}$ \citep{ds} in English.

For fair comparison, we use word2vec skip-gram embeddings openly available from the NLPL vector repository \footnote{\href{http://vectors.nlpl.eu/repository/}{http://vectors.nlpl.eu/repository.}} \citep{wordemb} and enhance token representations with multilingual BERT \cite{bert}, which has 12 transformer blocks, 12 attention heads, and 768 hidden units.
Our network weights are optimized with Adam and we also conduct Cosine Annealing Warm Restarts learning rate schedule \citep{cawr}. We fixed the word embeddings during training process. The char embedding size is set to 100. The dropout rate of embeddings and other network components are set to 0.4 and 0.3 respectively. 
We employ 4-layer BiLSTMs with the output size set to 400 and 2-layer for multi-hop reasoning with output size set to 768. The learning rate is 3e-5 and the batch size is 8. The hyperparameter $\alpha$ in Eq.\ref{alpha} is set to 0.25 (cf. Section \ref{search_alpha}). We use GeForce RTX 3090 to train our model for at most 100 epochs and choose the model with the highest SF1 score on the validation set to output results on the test set.


\subsection{Baselines}
We compare our proposed model with three state-of-the-art baselines which outperform other models in all datasets:
\paragraph{RACL-BERT}\citet{racl} propose a relation-aware collaborative learning framework for end2end sentiment analysis which models the interactive relations between each pair of sub-tasks (target extraction, expression extraction, sentiment classification). \citet{structured} reimplement the RACL as a baseline for SSA task in their work. 

\paragraph{Head-first and Head-final\footnote{\href{https://github.com/jerbarnes/sentiment_graphs}{https://github.com/jerbarnes/sentiment\_graphs.}}} \citet{structured} cast the structured sentiment analysis as a dependency parsing task and apply a reimplementation of the neural parser by \citet{biafparsing}, where the main architecture of the model is based on a biaffine classifier. The Head-first and Head final are two models with different setups in the parsing graph.

\subsection{Evaluation Metrics}
Following previous SOTA work \citep{structured}, we use the Span F1, Targeted F1 and two Sentiment Graph Metrics to measure the experimental results. 

In detail, Span F1 evaluates how well these models are able to identify the holders, targets, and expressions. Targeted F1 requires the exact extraction of the correct target, and the corresponding polarity. Sentiment Graph Metrics include two F1 score, Non-polar Sentiment Graph F1 (NSF1) and Sentiment Graph F1 (SF1), which aims to measure the overall performance of a model to capture the full sentiment graph (Figure \ref{fig1}(a)). 
For NSF1, each sentiment graph is a tuple of (holder, target, expression), while SF1 adds the polarity (holder, target, expression, polarity). A true positive is defined as an exact match at graph-level, weighting the overlap in predicted and gold spans for each element, averaged across all three spans.

Moreover, for ease of analysis, we add an Overall Span F1 score which evaluates how well these models are able to identify all three elements of a sentiment graph with token-level F1 score.

\begin{table}[t]
\resizebox{.99\columnwidth}{!}{
\begin{tabular}{@{}lccc@{}}
\toprule
                            & Span Overall F1 & Targeted F1 & SF1  \\ \midrule
\textbf{Ours(TGLS)}                 & \textbf{58.1}            & \textbf{38.1}     & \textbf{37.6} \\\hdashline
\textcolor{white}{pad}w/o \texttt{[CLS]}-related graph & 57.6            & 36.9     & 36.1 \\
\textcolor{white}{pad}w/o span graph              & 57.2            & 38.1     & 37.4 \\
\textcolor{white}{pad}w/o relation graph          & 57.7            & 38.0     & 36.1 \\
\textcolor{white}{pad}w/o vanilla GAT graph       & 57.8            & 37.6     & 36.5 \\\hdashline
\textcolor{white}{pad}w/o RoPE                    & 57.7            & 36.4     & 36.8 \\
\textcolor{white}{pad}w/o adaptive thresholding      & 56.0              & 36.3     & 35.2 \\ \bottomrule
\end{tabular}}
\caption{Experimental results of ablation studies.}
\label{ablation}
\end{table}

\begin{table}[t]
\centering
\resizebox{.99\columnwidth}{!}{
\begin{tabular}{@{}lccccc@{}}
\toprule
\multicolumn{1}{l}{} & \textbf{NoReC}$_\textsuperscript{Fine}$ & \textbf{MultiB}$_\textsuperscript{EU}$   & \textbf{MultiB}$_\textsuperscript{CA}$   & \textbf{MPQA }                  & \textbf{DS}$_\textsuperscript{Unis}$   \\ \midrule
Head-final           & 52.3  & 63.9 & 67.3 & 45.0 & 41.5 \\ \hdashline
TGLS model &&&&&\\
\textcolor{white}{pad}+parsing labels          & 54.2      & 65.4     & 67.5     &  44.7                      &  43.2    \\ 
\textcolor{white}{pad}+our labels               & \textbf{57.8}  & \textbf{68.7} & \textbf{70.1} & \textbf{46.1}                   & \textbf{45.7} \\ \bottomrule
\end{tabular}}
\caption{Experimental results of the relation extraction F1 score, where \textit{parsing labels} denote the dependency-parsing-based labels in head-final setting, \textit{our labels} denote the whole and essential labels.
}
\label{table2}
\end{table}

\subsection{Main Results}
In this section, we introduce the main experimental results compared with three state-of-the-art models RACL-BERT \citep{racl}, Head-first and Head-final models \citep{structured}.

Table \ref{table1} shows that in most cases our model performs better than other baselines in terms of the Span F1 metrics across all datasets. The average improvement ($\uparrow1.4$) in Overall Span F1 score proves the effectiveness of our model in span extraction.
Besides, there exists some significant improvements such as extracting holder on \textbf{DS}$_\textsuperscript{Unis}$ ($\uparrow$6.3) and extracting expression on \textbf{NoReC}$_\textsuperscript{Fine}$ ($\uparrow$4.7), but the extracting expression on \textbf{DS}$_\textsuperscript{Unis}$ ($\downarrow$2.9) are poor.  

As for the metric of Targeted F1, although the Head-first model performs well on \textbf{MPQA}, our TGLS model is obviously more robust as we achieves superior performance on other 4 datasets. There are also extremely significant improvements such as on \textbf{NoReC}$_\textsuperscript{Fine}$ ($\uparrow$6.2) and on \textbf{MultiB}$_\textsuperscript{CA}$ ($\uparrow$5.6), it proves the capacity of our model in exact prediction of target and the corresponding polar.

As for the Sentiment Graph metrics, which are important for comprehensively examining span, relation and polar predictions, our TGLS model achieves superior performance throughout all datasets in both NSF1 and SF1 score, especially on \textbf{NoReC}$_\textsuperscript{Fine}$ ($\uparrow$7.2 and $\uparrow$6.4). And the average improvement ($\uparrow$4.5) in SF1 score verifies the excellent ability of our model in the end-to-end sentiment tuple extraction. 

\begin{figure}[t]
\centering
\includegraphics[width=1\columnwidth]{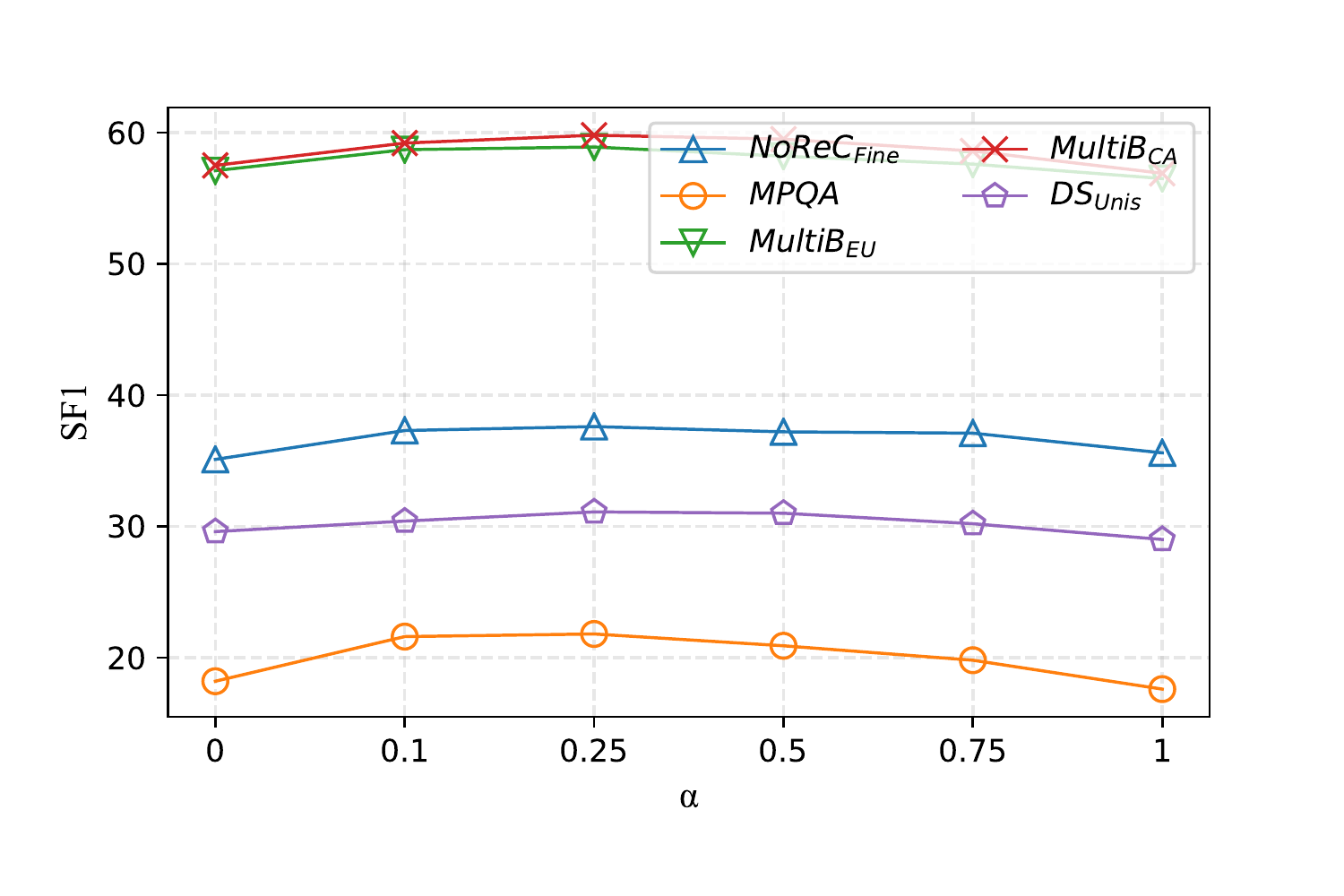}
\vspace{-8mm}
\caption{Experimental results (SF1 score) using different $\alpha$ to control the impact of the whole label prediction.}
\label{fig4}
\end{figure}

\begin{figure}[t]
\centering
\includegraphics[width=1\columnwidth]{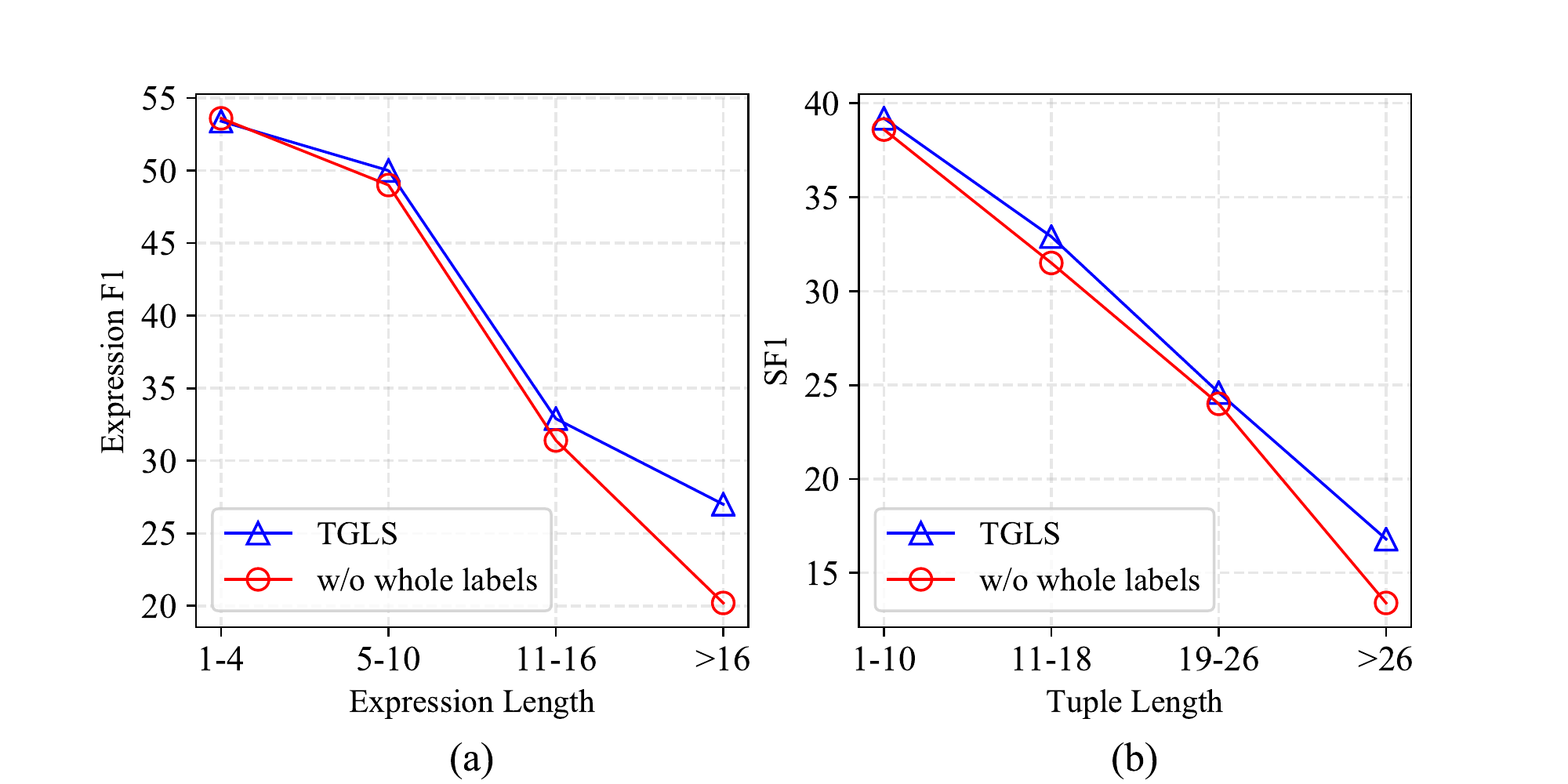}
\vspace{-4mm}
\caption{Analysis on the effect of the whole label set for long span identification. \textbf{(a)} Expression F1 scores regarding to different expression lengths. \textbf{(b)} SF1 scores regarding to different tuple lengths.
}
\label{fig5}
\end{figure} 

\subsection{Ablation Study}
In this section, we conduct extensive ablation studies on \textbf{NoReC}$_\textsuperscript{Fine}$ to better understand independent contributions of different components in terms of span overall F1, targeted F1 and SF1 scores.

Firstly, we remove each view of our graphs separately. As shown in Table \ref{ablation}, we observe that the \texttt{[CLS]}-related graph is effective in all three metrics which proves the importance of utilizing sentence-level global information. As we assumed, the span graph makes more contribution to the performance of span extraction (Span Overall F1) while the relation graph contributes more to end-to-end sentiment tuple extraction (SF1). And we also observe that the vanilla GAT graph makes considerable improvement in SF1 score.

Then, we test the effectiveness of the Rotary Position Embedding (RoPE) \cite{RoFormer}. The results in Table \ref{ablation} demonstrate that RoPE can make our model more sensitive to the relative positional information since it significantly improves the performance of exact target extraction (Targeted F1).

Last, we replace the adaptive threshold with fixed global threshold, and we observe that the performance drops drastically in all three metrics, it suggests that the adaptive thresholding mechanism is very crucial for our model since the flexibility can allow our model to selectively learn more useful information for SSA task from whole labels.

\section{Analysis}

In this section we perform a deeper analysis on the models in order to answer three research questions:

\subsection{Does our modeling strategy mitigate the label imbalance problem in span prediction and span relation prediction?}

Experimental results in Table \ref{table1} show that our model performs significantly better in the SF1 score, which to some extent proves that our model can ensure the efficiency of relation extraction. However, there lacks a metric to directly quantify the ability in relation extraction and it is still a worthy question to explore how much of the improvement comes from our new model and how much from our new labeling strategy?

To answer the question, we replace our labels with the dependency-parsing-based labels in head-final setting \citep{structured} and experiment on all datasets in terms of a new relation prediction metric, where a true positive is defined as any span pair that overlaps the gold span pair and has the same relation.
Table \ref{table2} shows that our new model achieves superior performance of relation prediction than the previous sota model \citep{structured}. Besides, with new labeling strategy, we can see that our model significantly improve the performance on all datasets compared with the model with replaced dependency-parsing-based labels.

\subsection{What is the appropriate value for the hyperparameter $\alpha$ in Eq. \ref{alpha}?
}
\label{search_alpha}
In this section,  we experiment on five datasets to heuristically search for the appropriate value of hyperparameter $\alpha$ (cf. Eq.(\ref{alpha})). Figure \ref{fig4} shows that all datasets achieve higher SF1 score with $\alpha$ between 0.1 and 0.5. We ended up fixing alpha to 0.25, since most datasets yield optimal results around this value. In addition, it is worth noting that when $\alpha$ is set to 0, which means that the whole labels are completely removed, the performance drops a lot, which once again proves the effectiveness of learning whole labels in the hidden layer.

\subsection{Is the whole label set helpful for long span identification?}



In this section,  we experiment on \textbf{NoReC}$_\textsuperscript{Fine}$ to further explore whether whole labels contribute to long span identification.
Figure \ref{fig5}(a) evaluates the Expression F1 scores regarding to different
expression lengths, we can find that whole labels helps most on those expressions with longer length.
In Figure \ref{fig5}(b), we also report the SF1 scores regarding to different distances, that is, from the leftmost token in a tuple to the rightmost token, which shows a similar conclusion.

\section{Conclusion}

In this paper, we propose a token graph model with a novel labeling strategy, consisting of the whole and essential label sets, to extract sentiment tuples for structured sentiment analysis.
Our model is capable of modeling both global and local token pair interactions by jointly predicting whole labels in the hidden layer and essential labels in the output layer. More importantly, our modeling strategy is able to alleviate the label imbalance problem when using token-graph-based approaches for SSA.
Experimental results show that our model overwhelmingly outperforms SOTA baselines and improves the performance of identifying the sentiment components with long spans.
We believe that our labeling strategy and model can be well extended to other structured prediction tasks.

\section*{Acknowledgements}
We thank all the reviewers for their insightful comments.
This work is supported by the National Natural Science Foundation of China (No. 62176187), the National Key Research and Development Program of China (No. 2017YFC1200500), the Research Foundation of Ministry of Education of China (No. 18JZD015), the Youth Fund for Humanities and Social Science Research of Ministry of Education of China (No. 22YJCZH064), the General Project of Natural Science Foundation of Hubei Province (No.2021CFB385).



\bibliography{anthology,custom}
\bibliographystyle{acl_natbib}

\end{CJK}
\end{document}